\def\year{2019}\relax
\documentclass[letterpaper]{article} 
\usepackage{aaai19}  
\usepackage{times}  
\usepackage{helvet}  
\usepackage{courier}  
\usepackage{url}  
\usepackage{graphicx}  
\usepackage{amsfonts}
\usepackage{wrapfig,float}
\usepackage{subfig}
\begin{document}
%
\title{Bootstrapping Conversational Agents With Weak Supervision}
\author{Neil Mallinar, Abhishek Shah, Rajendra Ugrani, Ayush Gupta, Manikandan Gurusankar,  
\vspace{0.5em} \\ \vspace{0.5em}
{\Large \bf Tin Kam Ho, Q. Vera Liao$^{1}$, Yunfeng Zhang$^{1}$, Rachel K.E. Bellamy$^{1}$, }\\
{\Large \bf Robert Yates, Chris Desmarais, Blake McGregor}\\
\\
IBM Watson, New York, NY 10003, USA \\
$^{1}$IBM Research AI, Yokrtown Heights, NY 10598, USA
\\
}

\maketitle
\begin{abstract}
Many conversational agents in the market today follow a standard bot development framework which requires training intent classifiers to recognize user input. The need to create a proper set of training examples is often the bottleneck in the development process. In many occasions agent developers have access to historical chat logs that can provide a good quantity as well as coverage of training examples. However, the cost of labeling them with tens to hundreds of intents often prohibits taking full advantage of these chat logs. In this paper, we present a framework called \textit{search, label, and propagate} (SLP) for bootstrapping intents from existing chat logs using weak supervision. The framework reduces hours to days of labeling effort down to minutes of work by using a search engine to find examples, then relies on a data programming approach to automatically expand the labels. We report on a user study that shows positive user feedback for this new approach to build conversational agents, and demonstrates the effectiveness of using data programming for auto-labeling. While the system is developed for training conversational agents, the framework has broader application in significantly reducing labeling effort for training text classifiers.
\end{abstract}
\section{Introduction}
Many conversational agents or chatbots nowadays are developed with a standard bot framework where a key step is to define \textit{intents} and build \textit{intent classifiers}. An intent in a conversational model is a concept that represents a high-level purpose of a set of semantically similar sentences for which the chatbot can provide the same response. Intents of user input are recognized by statistical classifiers trained with sample utterances.  For example, utterances like ``good morning'', ``hello'', ``hi'' are all for the intent of \textit{greeting}, and they can be used as training examples to train the classifier to recognize \textit{greeting}. This intent training step is supported by popular chatbot development platforms such as IBM Watson Assistant and Microsoft Azure Bot Service, but it is well recognized that high-quality training data is hard to obtain, resulting in sub-optimal intent recognition performance. A valuable resource is chat logs in relevant task domains, whether they are from conversations between users and human agents or collected from previous user interactions with a chatbot. 

However, using chat logs to bootstrap intents requires intensive labeling effort. Several characteristics of the chatbot development task make the conventional labeling process prohibitively expensive to apply. First, the intent classes are usually highly skewed, with a very small portion of positive examples present in the chat logs. So it would be expensive to obtain enough positive examples if labeling data in sequence. Second, a chatbot system usually has tens to hundreds of intents. Manually examining every piece of chat data and selecting one out of tens to hundreds of labels would be extremely challenging. For these reasons, chatbot development is often unable to take advantage of historical chat logs, but still largely relies on manual creation of user utterances as training data. 

In this work, we present a framework and build a system to drastically reduce the labor required in labeling chat logs, by harvesting a recently developed weak supervision technique called data programming~\cite{NIPS2016_6523}. Specifically, the system allows the chatbot developer or labeler to explicitly search training examples from chat logs for a given intent, then with minimum labeling input it uses the search queries for data programming to automatically propagate the label set. We call this framework SLP--search, label, and propagate. This framework offers several benefits: 1) it significantly reduces the labeling effort down to authoring search queries and providing a minimum number of labels to guide data programming; 2) it allows the labeler to focus on one intent at a time, thus making the labeling process easier to follow; 3) it can potentially support collaboration on labeling, by having multiple people authoring queries; 4) it can potentially make the management of relabeling process easier, as the labeler could easily edit or add queries and then have the system update the label set.

To demonstrate some of these benefits, we built a prototype system and conducted a user study to evaluate the training results with novice labelers with a merely 8-minute training session. By empirically studying the usage of a first-of-its-kind system, we also identified areas for future work to improve this new approach for developing chatbots, as well as the emerging area of weak supervision applications.

More importantly, our work represents an effort to apply weak supervision techniques to tackle a real world problem of training data bottleneck. Data programming allows labelers to express domain heuristics as labeling functions, which are programs that label subsets of the data, then automatically ``de-noises'' and propagates the labels. Beyond that, we propose to use a \textit{search engine} as a unified interface for authoring labeling functions, and use the search ranking to generate weak labels. Not only does it relieve users from explicitly programming labeling functions, but it also allows users to explore the dataset to formulate more queries. In the user study, we compared the training performance of applying data programming to merely using search to assist labeling of positive examples, which by itself is proposed to be a solution for labeling highly skewed data \cite{attenberg2010label}. We find that data programming improves the training performance and demonstrate its potentials in significantly reducing time and effort in creating training data. While SLP is developed for bootstrapping conversational agents, the framework can have broader application in significantly reducing labeling effort for text classifiers.

\section{Related Work}

\begin{figure}
\centering
\includegraphics[width=0.9\linewidth]{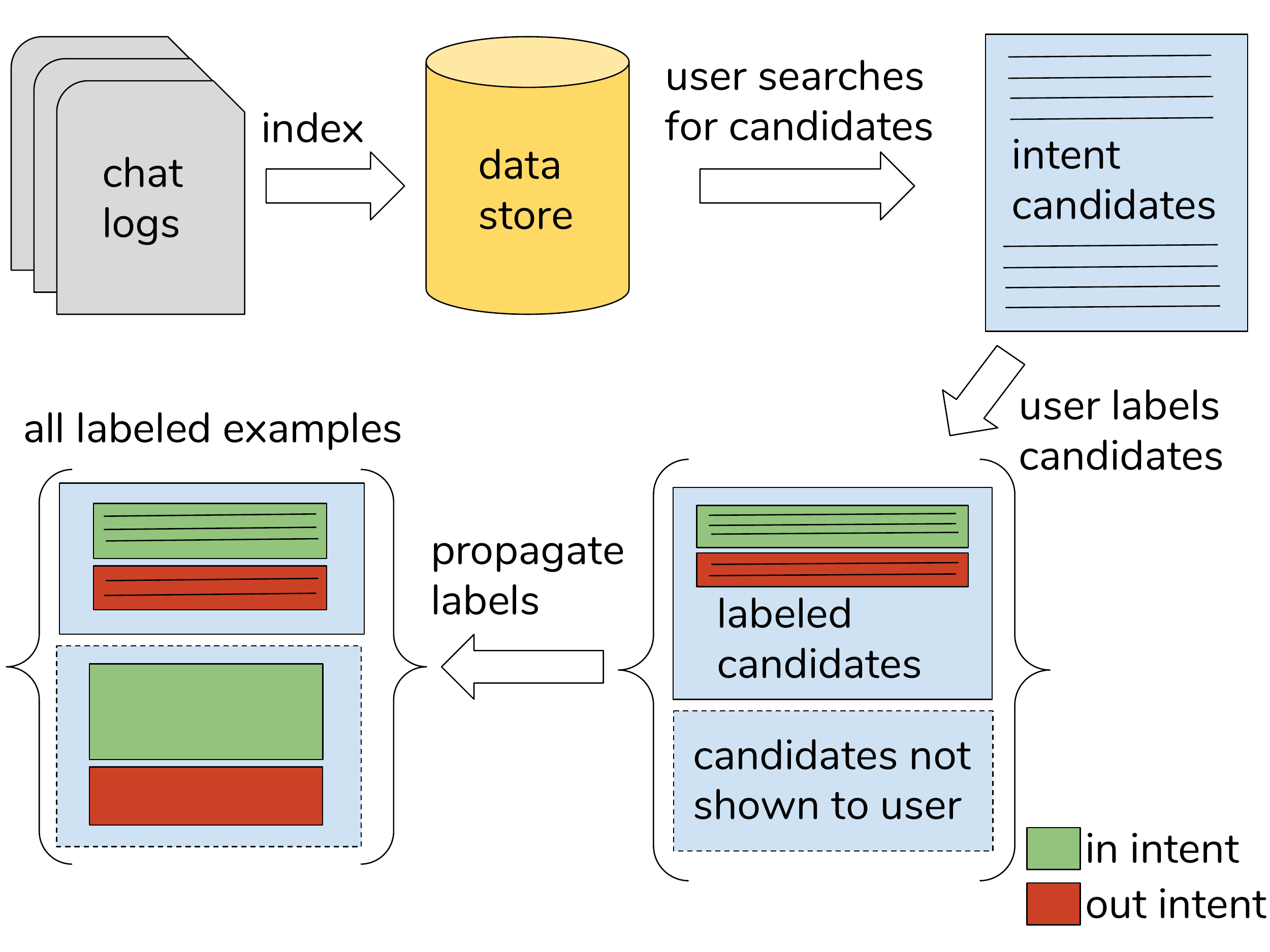}

\title{Overview of the Process}
\caption{Overview of the SLP process for generating a label set for an intent from chat logs.}
 \label{fig:framework}
\end{figure}

Adding intents is recognized as the main bottleneck in scaling--adding functionality to--conversational agents~\cite{williams2015rapidly}. Recent work focuses on two directions to improve the process. One is on bootstrapping--reusing available chat log data~\cite{goyal2018fast}--to rapidly expand the understanding and responding capabilities of agents. The other is to allow domain experts, who need not to be machine learning experts, to build intent models by working on model definition, labeling, and evaluation through user-friendly interfaces~\cite{williams2015rapidly}. Our work targets both directions by enabling an experts-in-the-loop approach to bootstrap intents.

In building conversational agents and other machine learning applications, the expensive cost to obtain sufficient and good-quality labeled data is a major obstacle. Given that, significant research effort has been made on reducing labeling effort, including work on semi-supervised learning~\cite{chapelle2009semi}, active learning~\cite{settles2012active}, and transfer learning~\cite{pan2010survey}. Our work is most relevant to the emerging area of weak supervision--approaches to obtain noisy but cost-efficient labels, especially the recently proposed data programming framework~\cite{NIPS2016_6523,ratner2017snorkel}. 

Weak supervision can be provided by various sources: by subject matter experts (SME) to provide higher-level, less precise heuristic rules, by cheap, low-quality crowdsourcing, or by taking advantage of external knowledge sources to heuristically align data points. The challenge is to combine weak supervision sources that may be overlapping or conflicting to increase the accuracy and coverage of the training set. Data programming allows programmatic creation of weak supervision rules in the form of labeling functions. It then builds a generative model of the data using the ensemble of labeling functions and the estimated dependency structure among them, which de-noises the resulted training set. The output is a set of probabilistic labels that can be used to train a discriminative model to generalize beyond the labeling functions and increase the label coverage.

While earlier work explored the idea of aggregating or modeling noisy labels from multiple sources with a semi-supervision approach~\cite{fujino2005hybrid,yan2016robust}, a key contribution of the data programming work is to provide a unified framework for programming heuristic rules. This is especially useful for soliciting domain specific heuristics from subject matter experts (SMEs). However, programming labeling functions can still be a burden for SMEs, who often have little programming experience. As shown in the user study of the data programming work~\cite{ratner2017snorkel}, it took hours for SMEs to learn and program labeling functions. A key contribution of our work is to explore using a unified interface for users to author labeling functions \textit{through interactions}. Unlike information extraction tasks explored in Ratner et al.'s work, we focus on text classification, where the forms of effective heuristic rules are more suitable to be entered via a unified interface. Specifically, we propose to use a search engine to solicit users to write search queries that retrieve training examples (both positive and negative ones), then automatically generate labeling functions based on the search queries. The novelty of this approach not only lies in the natural interaction for authoring labeling functions, but also in exploring using search ranking to generate weak labels.

The idea of using search to explicitly acquire training examples is relevant to guided learning~\cite{attenberg2010label}, an approach to reduce labeling effort under skewed classes, where the search approach is proved to be superior over labeling from uniform sampling and active learning. This is critical for the use case of bootstrapping conversational agents, as intents are fine-grained concepts and usually highly skewed in chat logs. We note that the same search interface can be used for guided learning, but our approach differs in that it takes the user's search queries to automatically expand the labeled set, and thus requires minimum labeling effort from the user. A contribution of our user study is to empirically compare the effectiveness of guided learning and our SLP framework.

\section{Method and System}

Our method works within the data programming framework by defining a set of labeling functions, each of which is equivalent to a weak classifier constructed with independent coverage to label a subset of the data. These labeling functions are built in a three step process: search, label, and propagate, where user input is only required for the first two steps. Specifically, the system generates a set of labeling functions based on the user's search queries and labels, and performs a structure learning process to model dependencies between them (allowing to relax the assumption that each weak classifier is independent). A generative model is trained to learn the parameters of the probability distributions induced by these labeling functions. Marginal probability values (in the case of categorical data, estimates of how likely a sample belongs to each of the classes) are calculated for each example in the corpus under the coverage of some labeling function(s).  Figure~\ref{fig:framework} shows the overview of the SLP process. After that, an optional step can utilize training a discriminative model on this set of examples using their relevant marginal probabilities as labels to further expand the label set, as in the original data programming framework.
\subsection{Search and Label}
We employ a ``search engine" framework to allow users to pull relevant examples out of provided corpora (in our case chat logs) which we call candidates. Our definition of ``search engine" is broad and there can be numerous methods by which relevant examples are retrieved, including but not limited to: lexical similarity, semantic similarity, strict entity/keyword matching, etc. The choice of method would impact the nature of the input (e.g., whether using keyword query or a full sentence query), as well as the nature of the pulled results (both precision and recall of the neighborhood retrieved would be affected by this choice). We proceed in this paper using the well-established and open source Elasticsearch engine for corpus indexing and searching. 

Elasticsearch allows for flexibility in user input, as one can construct a variety of queries ranging from simple to complex using the provided JSON-based query language. For the purpose of our study, we asked the user to search using short phrases and keywords, optionally providing boolean operators for moderately complex operations, or quotations to differentiate strict and fuzzy string matches. We opted for the standard Okapi-BM25 \cite{robertson2009probabilistic} ranking scheme to score candidate examples retrieved from chat logs by a query and we defined the neighborhood via a bound on the number of search results returned, being the top $N$. In a simulation environment set up to mock the user study design, we experimented with numerous settings of $N$, along with other parameters to be defined later. For simplicity, we will use $N=100$ as the default setting for the rest of the paper, however we believe that a dynamic setting based on the estimated size of the intent being modeled would perform better.

Once we established a neighborhood of size $N$ given an input query, we chose a subset of $k$ candidates from this neighborhood to display to the user for labeling. In the SLP framework, a small set of labeling input is needed from the user to serve two purposes: one is to provide information for propagation (e.g., most examples in this neighborhood are positive);  the other is to use them as ``strong labels'' to aid in the structure and generative model learning process. We note that the labeling interaction should be designed based on the method of propagation. With the propagation method we chose to implement for the user study (introduced in the next section), we parameterized the size of the subset-to-be-displayed by $k=10$, a value determined by user attention-span as well as the setting of $N$. For this set, we randomly sampled $\sim \frac{1}{3}k$ candidates from the bottom, middle, and top of the retrieved neighborhood, sorted by relevancy score. This allows the user to see a representative sample of the neighborhood that their decisions are propagated to.

\begin{figure}
\centering
\includegraphics[width=0.75\linewidth]{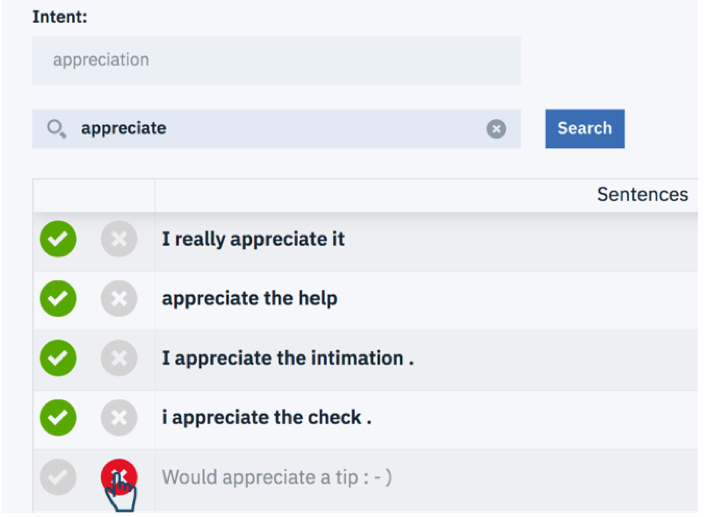}

\title{search interface}
 \vspace{-1em} 
\caption{Search and label interface of the prototype system}
\label{fig:search}
\end{figure}

The user labels requested are a simple ``in'' or ``out'' as to whether the given example belongs in the class currently being modeled, or not.  Optionally, a third state can be provided to allow a user to abstain from making a decision. Figure~\ref{fig:search} shows the interface of our prototype system for performing search and labeling.
\subsection{Propagate}
The propagation step of our method takes user labels on a small subset of candidates from a neighborhood and determines the best way to extend those labels to the entire retrieved neighborhood. While more sophisticated propagation methods can be used (which may require different designs of the labeling step), in the study we opted for a simple thresholded majority vote approach, as we believe that a sufficiently small setting of $N$ can pull highly precise neighborhoods from chat logs.

Let $k_{in}$ be the number of candidates marked in-intent by a user on the subset displayed. Similarly, let $k_{out}$ represent the number of candidates marked out-of-intent. For propagation, we first compare $\frac{k_{in}}{k}$ and $\frac{k_{out}}{k}$ to a threshold, $A$. In our application, we picked $A=0.6$. If either of the aforementioned ratios is above this threshold, we pick that label to propagate to the rest of the neighborhood. If neither falls above the chosen threshold we omit label propagation as this implies that the neighborhood is too noisy to meaningfully generalize to. In these cases, we do not utilize the unlabeled portion of the neighborhood in our learning process.

When the user finishes querying and all the propagations are complete, the learning portion of SLP is triggered using the user provided labels and their respective neighborhoods as inputs. 

\subsection{Learning with Weak Supervision}
The learning methodology in our system utilizes weak supervision techniques, in which (noisy) labels are assigned to a subset of data via a set of heuristic rules, also referred to as labeling functions. Labeling functions can range from keyword matching to using a noisy model trained on a small number of examples to distance-based metrics on vector embeddings. 

In our setting, we generate such labeling functions using the aforementioned Search and Label procedure, where each retrieved neighborhood is used as one weak labeling function. Given a set of $L$ labeling functions and $m$ candidate examples in our corpus, we assemble the user input into a label matrix, $\Lambda \in \mathbb{Z}^{L \times m}$, wherein each of the $m$ candidate examples is assigned a label in $\{-1, 0, 1\}$ by each labeling function. We allow for each labeling function to optionally abstain from making a decision if there is not enough information to make even a weak decision, using the label $0$ to represent abstention. While our example is for binary label sets it can be easily extended to categorical values, though we will continue to work with binary labels in this paper. 

We then learn the parameters of a generative model for the candidates included in the intent, as proposed by~\cite{NIPS2016_6523}, from our observed data, $\Lambda$. Specifically, the parameters of the model are $\alpha, \beta \in \mathbb{R}^L$ where $\alpha_i$ represents the likelihood of labeling a candidate correctly and $\beta_i$ is the likelihood that the $i^{\mathrm{th}}$ labeling function assigns a label (as opposed to abstains from labeling). 

We additionally learn dependencies between labeling functions through a structure learning process given by \cite{DBLP:journals/corr/BachHRR17}, by modeling the distribution in question as a factor graph. The most common dependencies are fixing and reinforcing. A fixing dependency is used to indicate that when two labeling functions assign disagreeing labels, one of them is correct over the other. A reinforcing dependency indicates that two labeling functions tend to both agree on the assigned label. 

After training the generative model we compute marginal probabilities per candidate example, where these values represent how likely it is for the example to be in-class. Additionally, any strong labels obtained during the Search and Label process are provided directly to the generative model as a single unified labeling function with a prior probability of $1.0$ of labeling correctly.

These steps working together would propagate user-provided labels to the unreviewed set of candidates by using search queries as labeling functions and a generative model to de-noise the propagated (weak) labels. These propagated labels can be used directly as training data (either as marginal probabilities or after conversion to binary labels based on a probability threshold), or as in the original data programming work, to train a discriminative model to further expand the label set to other unseen candidate examples. In the experiment, we use the user-provided and propagated labels, respectively, to train random forest models, and evaluate the results of applying the trained models to a set of held-out test data with ground truth. In other words, we focus on evaluating the training performance of propagated labels resulted from SLP, and comparing it to that of strong labels directly obtained from users without the propagation step.

\section{Experiments}
We conducted a user study with a real-world scenario of bootstrapping intents for a chatbot for an IT company, where chat logs between customers and technical support (human) agents are used to train the chatbot to perform similar customer service tasks. The study was task based, where participants were asked to use the SLP system to train three given intents. For each intent, a participant was given a description of the intent and had eight minutes to search and label. 
\begin{table*}[t]
\centering
\begin{minipage}{1\textwidth}
\begin{tabular}{p{2.2cm}p{1.8cm}p{1.8cm} p{1.8cm} p{1.8cm}p{1.8cm} p{1.8cm} p{1.8cm}} \hline 
 & \textbf{N (query)} &\textbf{N (label)} & \textbf{accuracy} & \textbf{precision(+)} & \textbf{precision(-)} & \textbf{recall(+)} & \textbf{recall(-)} \\ \hline
(1) Label-only & 4.54 (1.79) & 77.42 (28.98) & 0.507 (0.408) & 0.041 (0.033)& 0.940 (0.201) & \textbf{0.550} (0.391) & 0.505 (0.425) \\
(2) SLP: strong & 9.09 (2.56) & 78.78 (20.16) & 0.781 (0.264) & 0.046 (0.030) & 0.980 (0.012) & 0.297 (0.255) & 0.794 (0.273)\\
(3) SLP: weak & - &401.57 (149.68)& \textbf{0.963} (0.039) & \textbf{0.190} (0.211)& \textbf{0.980} (0.012) & 0.153 (0.161) & \textbf{0.982} (0.038) \\ \hline
\end{tabular}

\vspace{-0.5em}
\end{minipage}
\caption{Mean statistics for models (standard deviation in the parentheses). + (-) means positive (negative) class. }
\label{tab:result}
\end{table*}


\begin{figure*}[t]
\centering
\begin{minipage}{1\textwidth}
\includegraphics[width=1.0\linewidth]{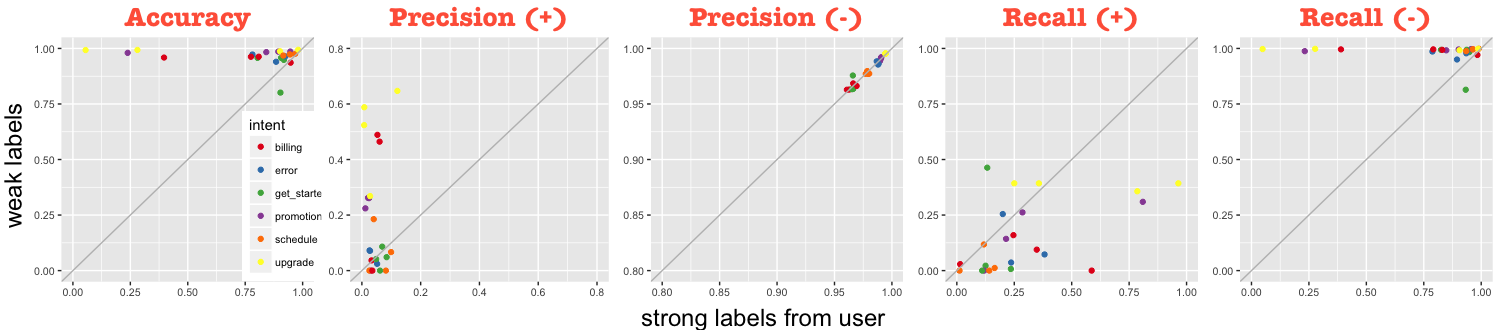}
\vspace{-2.0em}
\caption{Performances of models trained using strict labels versus weak labels in the SLP condition. + (-) means positive (negative) class.}
 \label{fig:resultplot}

\vspace{-1em}
\end{minipage}
\end{figure*}

\subsection{SLP v.s. Label-Only}
A key idea of our SLP framework is to rely on using search ranking as labeling functions of data programming to automatically propagate the labels. A focus of the user study is to evaluate the effectiveness of the propagation component. In fact, without data programming, the system can be used as \textit{guided learning}~\cite{attenberg2010label} by assisting labelers to actively search for positive examples to label. This approach has been proved to be effective in dealing with skewed classes. Therefore, we conducted an A/B testing experiment to compare the training performances of using the SLP framework versus using the system as guided learning (label-only).  

We randomly assigned participants to either the SLP or label-only conditions. In practice, they would be using the same interface but have different understanding on how the system works. For those using SLP framework, they should understand that the goal is to create as many search queries as possible for propagation, and only need to label a small number of examples. For those doing label-only, they should understand that the goal is to search for positive examples and label as many as possible. We reflected these differences in the task instructions received by participants in the two conditions. In addition, we reinforced the difference by showing only the representative 10 search results for those in the SLP condition (minimum labeling effort), while providing all top 100 search results with pagination for those in the label-only condition and suggesting them to label at least 20 examples for each search.

\subsection{Dataset, task and experiment procedure}
To simulate a real-world task, we obtained proprietary chat logs of customer service from the IT company. The corpus for bootstrapping intents consists of 40.8k utterances from customers after we excluded those with excessive length over 204 characters. For test data, we randomly sampled 3700 utterances and manually labeled them with regard to the intents used in the experiment.

We defined 6 intents that are frequent in this corpus and common for a customer service chatbot. Examples of these intents include: ``\textit{schedule}''--a customer requests to schedule a phone call or meeting with an agent; ``\textit{promotion}''--a customer inquires about getting or using promotional offers.

We recruited 16 participants from the IT company who are familiar with the products. They learned about basic concepts of training a chatbot before coming to the study. We randomly assigned half of them to the SLP condition, and the other half to the label-only condition. Each participant was given the task instruction and a training task to get familiar with the tool. They were then given 3 intents, randomly selected from the 6 intents we designed, to work on. They were timed for 8 minutes for each intent training task. After the experiment, each participant was interviewed for 10-15 minutes to gather feedback.

\section{Results}
In this section, we first compare the training performance of using weak labels generated by SLP and manual labels with assisted search (label-only). For the SLP condition, we also look into the training performance of directly using strong labels provided users, to further understand the effect of propagation. We then discuss observed user behavior and user feedback in using the SLP system, and their implications for future work. 

\subsection{Model performance}
As shown in Table~\ref{tab:result}, on average, for each intent task, participants in the SLP condition created 9.09 queries and provided 78.78 strong labels. The SLP framework then propagated to 401.57 labels. In the label-only condition, participants performed significantly less search (4.54 queries on average).  The strong labels in the label-only or the SLP processes are used to train random forest classifiers (lines (1) and (2)) and the propagated labels with marginal probabilities are used to train random forest regression models (line (3)).  The features are from a TF-IDF vectorization of the examples. All models are then applied to the held-out test set from which the statistics are calculated.
Comparing the training performance between using SLP (weak labels) and label-only, there is significant improvement in all measures with the only exception of recall on the positive class. In Figure~\ref{fig:resultplot}, we plot the performance of individual training tasks in the SLP condition using weak labels versus using strong labels, from which the weak labels are generated. We conclude that the propagation component significantly improves accuracy, precision on the positive class, and recall on the negative class, while paying the price of reducing recall on the positive class for some cases. 

This means that the SLP framework makes the classifier ``more strict''. For the specific task of training a chatbot, precision is often more important as it is preferable for a chatbot to acknowledge it does not recognize a user input rather than provide a wrong answer. It is notable that in Figure~\ref{fig:resultplot}, many tasks are significantly improved on the precision measure. We observe that this improvement is more common for the following three intents: \textit{promotion}, \textit{upgrade} and \textit{billing}. In the post-study interview, these intents were consistently reported to be ``narrower'', where the set of keywords (heuristic rules) were easier to recall and more constrained, in contrast to intents such as \textit{get started} where positive examples tend to take more diverse forms. It is plausible that the SLP framework is especially useful for training more well-defined classes, or when the labelers are highly familiar with the domain and the corpus. 

Meanwhile, we observe that the decrease in recall on the positive class happens already when comparing the performance of using the strong labels from the SLP condition to those from the label-only condition, where the difference is that users performed more search and labeled less per query in the former. We examined a few cases with particularly low recall, and observed a tendency for these users to add queries by rephrasing previous ones with more specific keywords (e.g., ``meeting'', ``schedule meeting'', ``schedule meeting time''). This could have led to more strict training data and is not an effective way to use the SLP system. In the future work, we will explore system functionality that helps users avoid this kind of behavior and author more effective and diverse search queries.

Lastly, we note that although some of the performance measures are relatively low, these are results from novice labelers (with only shallow knowledge of the domain) using the system for merely 8 minutes. We expect the performance measures to be significantly enhanced in the usage by actual chatbot developers who would be more familiar with the chat logs and thus can create more effective queries. Nevertheless, the user study highlights the evident improvement in applying the SLP framework.

\subsection{User feedback}


We surveyed participants with the System Usability Scale (SUS), and on average a score of 4.1 (out of 5) was reported, showing positive user feedback on the prototype system. From the user interview, we identified the following themes of user needs:
\begin{itemize}
\item Guidance on creating effective queries: different from the conventional usage of a search engine--i.e., finding the single best answer, the use of search in a data programming context should target high coverage~\cite{ratner2017snorkel} without over-sacrificing the precision in retrieving examples. Users desire to have guidance on how to optimize for precision, coverage, and bias between positive/negative examples in creating queries. Future work should explore providing such guidelines.
\item Support data exploration: users expressed difficulty in coming up with queries sometimes due to unfamiliarity with the corpus. Even for a chatbot developer familiar with the domain, the acquired corpora may differ in the types of inquiries and vocabularies used. Future work can provide functionalities for users to explore the dataset.
\item Feedback and progress tracking: users desire to see how each query or labeling function impacts the results with immediate feedback. The feedback can help users actively adjust the labeling functions they provided. Moreover, given that chatbot development requires training tens to hundreds of intents, it is critical to be able to make a decision on finishing training one intent and moving on to the next. Future work may have to explore metrics to provide fast feedback for weak supervision. 
\item Support evolving classes: training a classifier is often an evolving process as the labeler see more data points and refine the boundaries. Moreover, chatbot development often requires revising the intents as end users' behaviors or needs evolve. A potential benefit of the SLP framework is that it can make the re-labeling process easier--one only needs to update the queries and has the system to automatically re-generate the labels. Future versions of the system should support reviewing of the query history. 
\end{itemize}

In summary, we demonstrate that our SLP framework can significantly expand the training set and improve the training performance. It is especially helpful for improving the precision and thus creating higher-quality intents. Further, we point out the potential problems in creating ineffective queries that may harm the performance.  We also gathered user feedback to inform future work for the emerging applications of bootstrapping conversational agents, and more broadly training text classifiers, using weak supervision. 
\bibliographystyle{aaai}
\bibliography{bibliography}

\end{document}